\documentclass[11pt]{article}

\usepackage{dialogue}

\usepackage[most]{tcolorbox}

\usepackage{ifxetex}
\ifxetex
    \usepackage{fontspec}
    \setromanfont{Times New Roman}
\else
  \usepackage{cmap}
  \usepackage[T2A,T1]{fontenc}
  \usepackage[utf8]{inputenc}
  \usepackage{times}
  \usepackage{latexsym}
  \DeclareFontFamilySubstitution{T2A}{\familydefault}{cmr}
\fi

\usepackage[russian,british]{babel}
\usepackage{url}
\usepackage{pgf}

\usepackage{covington} %% comment it out if you do not require this option

\usepackage{hyperref}

\dialogfinalcopy % Uncomment this line for the final submission

\title{Can LLM Teams Play What? Where? When?}

\author{Anastasia Kotelnikova \\
  Vyatka State University \\
  Kirov, Russia \\
  {\tt kotelnikova.av@gmail.com} \\\And
  Viktor Byzov \\
  Vyatka State University \\
  Kirov, Russia \\
  {\tt vbyzov@yandex.ru} \\\AND
  Maria Dolzhenkova \\
  Vyatka State University \\
  Kirov, Russia \\
  {\tt maryd@vyatsu.ru} \\\And
  Evgeny Kotelnikov \\
  European University at St. Petersburg \\
  St. Petersburg, Russia \\
  {\tt kotelnikov.ev@gmail.com} \\
  }

\date{}

\begin{document}
\maketitle
\begin{abstract}

Large language models (LLMs) remain limited on tasks requiring indirect reasoning, cultural knowledge, and coordinated hypothesis testing. We investigate whether team-based interaction improves LLM performance in What? Where? When? (ChGK), a quiz game designed to reward collective reasoning. We introduce three team strategies: Voting, Silent Team (the captain observes final answers), and Talkative Team (the captain observes both answers and rationales). To minimize data leakage, we evaluate these strategies on a dataset consisting of 572 ChGK questions released in 2025.

Using six recent large-scale open models, we show that team-based strategies outperform single-model baselines, yielding gains of up to 20 percentage points in accuracy. The best team achieves 44.23\% accuracy, and approaches human team performance on questions with available human statistics. Analysis of inter-model diversity reveals that disagreement strongly predicts lower accuracy, but explanatory communication substantially mitigates performance drops. We further examine captain behavior and find no evidence of self-preference bias; access to peer rationales improves captain judgments.

Overall, LLM teams function primarily as answer selection and error-filtering mechanisms rather than generators of novel solutions. Our findings highlight the importance of interaction and suggest adaptive strategies as a promising direction for multi-agent systems.

  
  \textbf{DOI:}
\end{abstract}

\selectlanguage{russian}
\begin{center}
  \russiantitle{Могут ли команды больших языковых моделей \\ 
  играть в "Что? Где Когда?"}

\end{center}

\begin{abstract}
    Большие языковые модели (LLM) по-прежнему испытывают ограничения при решении задач, требующих неявных рассуждений, культурных знаний и координированной проверки гипотез. В работе исследуется, улучшает ли моделирование командного взаимодействия результаты LLM в игре «Что? Где? Когда?». Мы рассматриваем три стратегии организации LLM в команду: «голосование», «молчаливая команда» (капитан видит только ответы членов команды) и «разговорчивая команда» (капитан видит и ответы, и рассуждения). Чтобы минимизировать утечки данных, мы оцениваем эти стратегии на датасете, состоящем из 572 вопросов, выложенных в 2025 году.

    Используя шесть современных открытых LLM, мы показываем, что командные стратегии превосходят одиночные модели, обеспечивая прирост точности до 20 процентных пунктов. Наиболее успешная команда достигает точности 44,23\% и приближается к результатам человеческих команд на вопросах, для которых доступны статистические данные ответов людей. Анализ разнообразия ответов показывает, что рост расхождений между моделями связан со снижением точности, однако обмен рассуждениями смягчает падение качества. Также мы исследуем поведение капитана и не обнаруживаем эффекта предпочтения собственного ответа; доступ к рассуждениям членов команды повышает качество решений капитана.

    В целом, команды LLM выступают прежде всего как механизм отбора ответов и фильтрации ошибок, а не источник принципиально новых решений. Наши результаты подтверждают значимость взаимодействия между моделями и перспективность адаптивных стратегий для многоагентных систем.
  
  \textbf{Ключевые слова:} Большие языковые модели, викторины, многоагентные системы, коллективный интеллект, LLM в роли судьи
\end{abstract}

\selectlanguage{british}

\section{Introduction}
\label{introduction}

Even strong language models struggle with indirect cultural, metaphorical, or multi-step reasoning. A potential solution is team-based inference: multiple models answer independently, and a designated captain model aggregates their responses into a final decision.

Consider the following question from a quiz competition:

\begin{examples}
\item "One of Adriano Celentano’s songs consists of meaningless words imitating the sound of English. In an interview, he stated that the song reflects social fragmentation and mentioned a city. Which city?"
\end{examples}

In a six-model team simulation, four models proposed \textit{New York}, one suggested \textit{Milan}, and only one produced the correct answer, \textit{Babylon}. After reviewing all responses, the captain revised its initial prediction and selected \textit{Babylon}, recognizing the reference to the “Babylonian confusion of tongues.” This example illustrates how minority insights can become decisive when properly aggregated.

We evaluate collective reasoning in the context of \textit{What? Where? When?} (Russian: \textit{Chto? Gde? Kogda?}; \textit{ChGK}), a Russian team-based intellectual game characterized by riddle-like questions requiring indirect inference, linguistic sensitivity, and cultural knowledge.\footnote{\url{https://en.wikipedia.org/wiki/What\%3F\_Where\%3F\_When\%3F}} Unlike standard QA benchmarks focused on factual retrieval or localized reasoning \cite{ni2025}, ChGK questions are designed for collaborative problem solving, encouraging hypothesis comparison and correction of misleading interpretations \cite{Foster2025}. This makes ChGK a natural benchmark for collective intelligence.

Single LLMs exhibit well-documented limitations in complex reasoning tasks, including hallucinations \cite{alansari2025}, overconfidence \cite{guan2025}, narrow reasoning trajectories \cite{zheng2025}, and limited self-criticism \cite{stechly2025}. While methods such as chain-of-thought prompting \cite{zhu2025chain}, self-reflection \cite{zhu2025emergence}, and iterative refinement \cite{xue2025} improve internal reasoning, they remain constrained by the biases of a single model.

Recent work has therefore shifted toward ensemble \cite{bujnowski2025} and multi-agent approaches \cite{pitre2025}, where multiple models exchange information and aggregate predictions. This paradigm aligns naturally with ChGK-style tasks, which reward diversity of viewpoints and penalize premature convergence on plausible but incorrect answers. Human ChGK teams provide a useful analogue: they distribute cognitive roles, compare competing hypotheses, and sometimes rely on minority insights supported by key evidence.

We investigate whether explicit modeling of team interaction improves LLM performance on ChGK questions. We compare three team strategies: \textit{Voting} (majority aggregation), \textit{Silent Team} (captain observes final answers only), and \textit{Talkative Team} (captain observes answers and intermediate reasoning). These configurations disentangle the roles of diversity, explanation, and coordination in collective decision-making.

Our contributions are as follows:

\begin{itemize}

    \item We introduce three team-based interaction paradigms capturing different levels of information sharing.

    \item We construct a new evaluation dataset of 572 ChGK questions from 2025, designed to minimize data leakage.

    \item Using six recent large-scale open models, we show that team-based strategies consistently outperform single-model baselines, with gains of up to 20 percentage points in accuracy.

    \item We analyze disagreement and communication effects, showing that explanatory sharing is especially beneficial under high uncertainty.

    \item We examine captain decision behavior and demonstrate that access to peer rationales improves confidence calibration and reliability.

\end{itemize}

\section{Previous Work}
\label{previous_work}

\subsection{Ensemble and Multi-Agent Approaches for LLM-based Question Answering}

Recent work shows that combining multiple LLMs through ensembling or structured multi-agent interaction can substantially improve question answering performance.

Bujnowski et al. \cite{bujnowski2025} proposed a heterogeneous LLM ensemble with confidence-aware voting and arbitration, achieving strong results on tabular QA. In the medical domain, Yang et al. \cite{yang2025} demonstrated that question-adaptive weighting of complementary models outperforms uniform aggregation. Lu et al. \cite{lu2025} showed that diversity-aware ensembles consistently surpass single models, especially on complex reasoning tasks, while even simple majority voting can be effective in multimodal settings \cite{nguyen2025}.

Beyond static aggregation, interactive multi-agent frameworks have been introduced. Pitre et al. \cite{pitre2025} proposed a debate-based system in which agents iteratively exchange answers, explanations, and confidence estimates, leading to consistent improvements over single-agent and standard ensemble baselines.

Overall, prior work highlights the importance of diversity, adaptive aggregation, and structured interaction for improving the reliability and accuracy of LLM-based QA systems.

\subsection{Datasets}

Large quiz-style datasets are commonly used to evaluate QA systems and LLMs. Widely adopted resources include the \textbf{Jeopardy! clue dataset}\footnote{\url{https://github.com/jwolle1/jeopardy\_clue\_dataset}}, \textbf{TriviaQA} \cite{joshi2017}, \textbf{SearchQA} \cite{dunn2017}, and \textbf{QANTA} \cite{rodriguez2021}, which test knowledge retrieval, evidence-based reasoning, and incremental inference in quiz-like formats.

For Russian-language quiz games, the largest public resource is \textbf{Russian Jeopardy!} \cite{mikhalkova2022}, derived from db.chgk.info. The curated \textbf{CheGeKa} subset is included in the TAPE benchmark for few-shot Russian language understanding \cite{taktasheva2022}. More recently, \cite{kuznetsova2025} released a dataset of 2,600 ChGK questions from the IQ Game platform (2018–2025) for evaluating open-source LLMs.

A key challenge of large quiz archives is data leakage, as many questions have been publicly available for years and may appear in pretraining corpora, particularly affecting closed-book evaluation. To mitigate this issue, we construct a new dataset consisting exclusively of questions collected from the IQ Game platform\footnote{\url{https://iqga.me}} in 2025. Restricting evaluation to recent and previously unused material provides a more reliable estimate of genuine generalization and reduces the risk that model performance is driven by memorization of widely circulated quiz content.

\section{Methodology}
\label{methodology}

This section describes the experimental framework used in our study. We first describe the team strategies, which form the main methodological component of our approach, and then present the dataset, the models, and the evaluation protocol used in the experiments.

\subsection{Team Strategies}

We compare three strategies for combining the outputs of six LLMs, one of which acts as the captain.

\paragraph{Voting.}

In the Voting strategy, all six models answer the question independently (see Appendix~\ref{app:prompt_individual_answer}). 
Since answers that mean the same thing may differ in wording, we use \texttt{Gemini-2.5-flash} to normalize and group semantically equivalent responses (Appendix~\ref{app:prompt_gemini_normalize_answers}). 

After grouping, the answer supported by the largest number of models is selected. In the case of a tie, the captain acts only as a tie-breaker: if the captain’s original answer is among the tied options, it is selected; otherwise, one of the tied answers is chosen at random. As a result, Voting performance may differ depending on which model is designated as the captain.

\paragraph{Silent Team.}

In this setting, the captain receives the six answer variants and must decide on the final answer (Appendix~\ref{app:prompt_silent_team}). 
The captain does not know which answer was its own.
The captain may select one of the proposed answers or generate a new one if none appears correct.

\paragraph{Talkative Team.}

The Talkative Team extends the Silent Team setup by also providing the captain with short reasoning statements produced by each model. 
Thus, the captain sees both the proposed answers and brief explanations of how they were obtained.
Apart from this additional information, the procedure is identical to the Silent Team strategy (Appendix~\ref{app:prompt_talkative_team}).

\subsection{Dataset}
\label{dataset}

Our dataset consists of 572 ChGK questions collected from the IQ Game platform in 2025. All experiments were conducted in Russian, which is the original language of the questions. English translations shown in the paper are provided for readability only.
All questions are text-only and do not include images or other multimedia content.

Each question is paired with a canonical answer, a set of accepted alternatives, and an explanatory comment outlining the intended reasoning. 
When available, we also include human answer statistics showing how often the question was solved correctly by human teams. Since this information is missing for 133 questions, analyses involving human statistics are conducted on the remaining 439 questions.

\begin{figure}[t]
  \centering
  \includegraphics[width=0.8\linewidth]{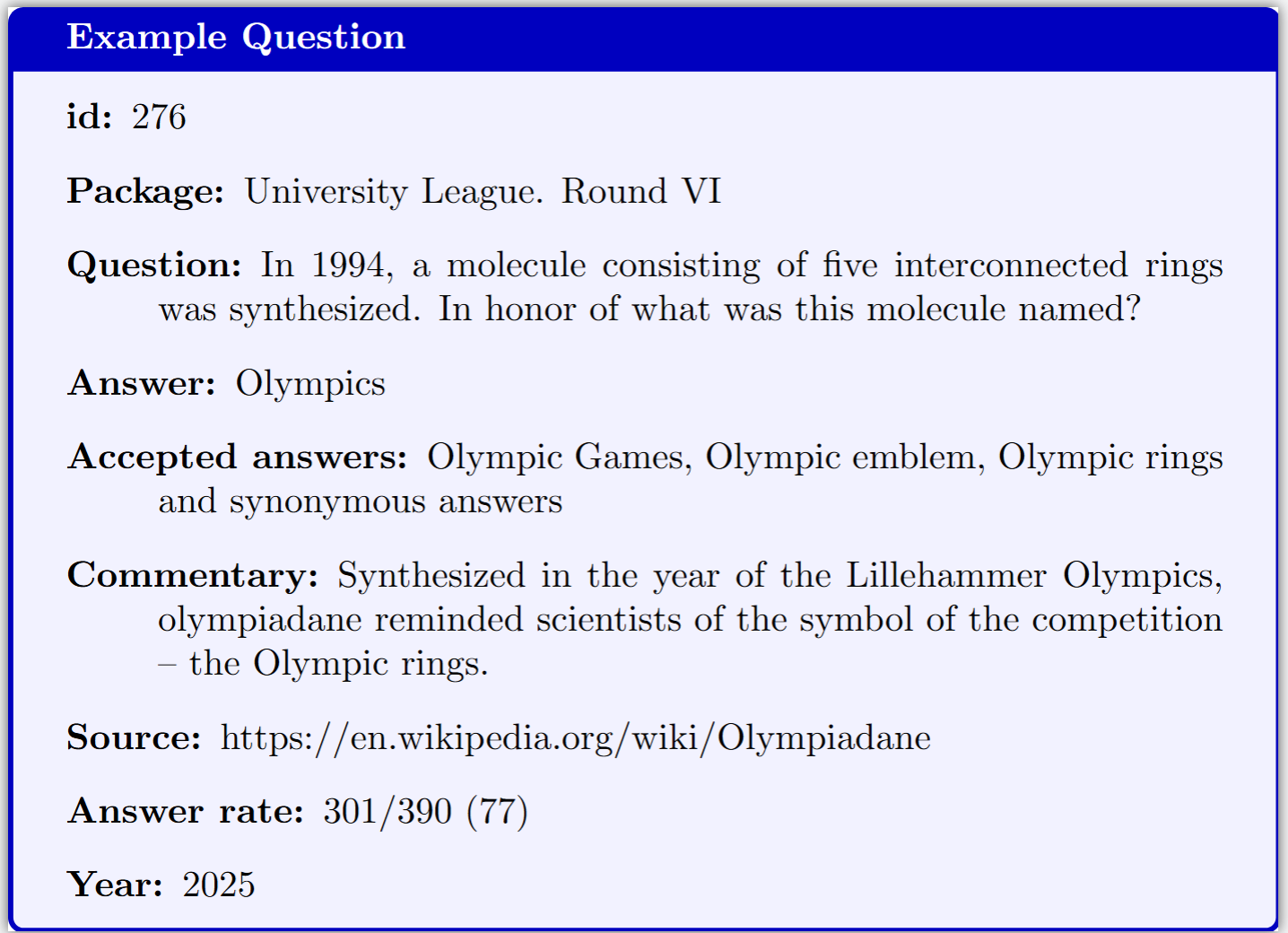}
  \caption{English translation of an illustrative quiz question from the original Russian dataset.}
  \label{fig:illustrative_example}
\end{figure}

An example question is presented in Figure~\ref{fig:illustrative_example}.

\subsection{Models}
\label{models}

Our team consists of six recent open-source LLMs accessed via public APIs. 
All models are large-scale Mixture-of-Experts systems released in 2025.

\paragraph{Qwen Family.}
\begin{itemize}
    \item \textbf{Qwen3-235B-A22B} (April 2025): 235B total parameters (22B active), general-purpose instruction and reasoning model.
    \item \textbf{Qwen3-235B-A22B-Thinking-2507} (July 2025): reasoning-oriented variant of the same architecture.
\end{itemize}

\paragraph{DeepSeek Family.}
\begin{itemize}
    \item \textbf{DeepSeek-V3.2} (December 2025): 671B total parameters (37B active), general-purpose model.
    \item \textbf{DeepSeek-R1-0528} (May 2025): 685B total parameters (37B active), optimized for reasoning.
\end{itemize}

\paragraph{Kimi Family.}
\begin{itemize}
    \item \textbf{Kimi-K2-Instruct-0905} (September 2025): ~1T total parameters (32B active), optimized for long-context instruction following.
    \item \textbf{Kimi-K2-Thinking} (November 2025): reasoning-focused variant of the same architecture.
\end{itemize}

\medskip

All models are used in their publicly available inference configurations without additional fine-tuning. 
We apply default decoding settings with temperature fixed to zero to ensure deterministic outputs.

This heterogeneous selection allows us to form a diverse team combining strengths in instruction following, long-context processing, and reasoning.

For each question, a model may generate up to five attempts. These retries are used only to handle occasional formatting errors or generation failures. If no valid answer is produced after five attempts, the question is marked as unanswered and counted as incorrect.

\subsection{Evaluation Protocol}
\label{evaluation}

We evaluate answers in two stages: 
(1) automatic string matching and 
(2) LLM-based verification for unresolved cases.

\paragraph{Stage 1: Automatic Matching.}

All answers are first preprocessed: we convert text to lowercase, remove diacritics, and apply lemmatization when specified in the dataset guidelines. 
If the question allows free word order, tokens are reordered before comparison.

The processed answer is then compared against the canonical answer and the set of accepted alternatives. 
If a match is found, the answer is marked as correct and no further checks are performed.

\paragraph{Stage 2: LLM-as-a-Judge.}

Answers not resolved at the first stage are evaluated by a judge model. 
We use \texttt{Gemini-2.5-flash}, following \cite{kuznetsova2025}, where it showed strong performance in answer verification at relatively low cost.

For each case, the judge assigns a binary label (correct or incorrect) based on the reference answers and their accepted variants. 
The evaluation prompt is provided in Appendix~\ref{app:prompt_llm_as_a_judge}.

\paragraph{Metric.}

Performance is measured using accuracy, defined as the proportion of correct answers among all evaluated instances.

\section{Results}
\label{sec:results}

Table~\ref{tab:AccuracyResults} reports accuracy for individual models and team-based strategies.

\begin{table}[t]
\begin{center}
\begin{tabular}{|l|r|rrr|}
\hline \bf Model & \bf Single Model & \bf Voting & \bf Silent Team & \bf Talkative Team \\ \hline
Qwen3-235B-A22B & 33.39 & 42.31 & 40.03 & 42.31 \\
Qwen3-235B-A22B-Thinking-2507 & \bf 37.41 & 43.01 & 42.48 & \bf 44.23 \\ \hline
DeepSeek-V3.2 & 35.14 & 42.83 & 36.19 & 41.61 \\
DeepSeek-R1-0528 & 29.20 & 40.91 & 37.24 & 39.51 \\ \hline
Kimi-K2-Instruct-0905 & 19.76 & 41.08 & 35.66 & 37.24 \\
Kimi-K2-Thinking & 30.77 & \bf 43.36 & \bf 42.13 & 37.06 \\
\hline
\end{tabular}
\end{center}
\caption{Accuracy of individual models and team-based strategies. 
Best results in each column are shown in \textbf{bold}.}
\label{tab:AccuracyResults} 
\end{table}

\subsection{Single-Model Performance}

The strongest standalone result is achieved by \texttt{Qwen3-235B-A22B-Thinking} (37.41\%), followed by \texttt{DeepSeek-V3.2} (35.14\%).

Within the Qwen and Kimi families, reasoning-oriented variants outperform their instruction-focused counterparts. 
For example, \texttt{Qwen3-235B-A22B-Thinking} exceeds \texttt{Qwen3-235B-A22B} by 4.02 p.p., and \texttt{Kimi-K2-Thinking} surpasses \texttt{Kimi-K2-Instruct} by 11.01 p.p.

This pattern does not hold for DeepSeek: \texttt{DeepSeek-V3.2} outperforms the reasoning-focused \texttt{DeepSeek-R1} (35.14\% vs.\ 29.20\%). 
This suggests that overall scale, training setup, and release maturity may be as important as explicit reasoning specialization.

\subsection{Impact of Team-Based Aggregation}

All team strategies substantially outperform single models, with gains ranging from roughly 8 to 20 percentage points.

The largest improvements occur for weaker baselines. 
For instance, \texttt{Kimi-K2-Instruct} improves from 19.76\% individually to 41.08\% under voting. 
These results show that aggregation effectively compensates for individual model limitations.

To estimate the upper bound of team performance, we also compute the skyline accuracy, i.e., the proportion of questions for which at least one model in the six-model pool produced a correct answer. Out of 572 questions, at least one model answered correctly in 341 cases, corresponding to 59.6\% accuracy. This result indicates a substantial gap between the current aggregation strategies and the maximal achievable performance obtainable through perfect answer selection.

\subsection{Comparison of Team Strategies}

Voting and captain-based approaches perform similarly overall, and no strategy consistently dominates.

The Talkative Team achieves the best results for both Qwen models, with \texttt{Qwen3-235B-A22B-Thinking} reaching the highest overall accuracy (44.23\%). For \texttt{Kimi-K2-Thinking}, Silent Team achieves 42.13\%, but Voting performs slightly better, reaching 43.36\% and yielding the strongest result for this model and among all Voting-based teams.

Overall, simple voting remains competitive, while captain-based strategies can offer additional gains in certain configurations.

\subsection{Model Families and Reasoning Orientation}

Qwen and DeepSeek models show relatively stable performance across settings, whereas the Kimi family exhibits greater variability between instruction and reasoning variants.

Reasoning-oriented models provide stronger single-model baselines, particularly in the Qwen and Kimi families. 
However, this advantage becomes smaller under team aggregation, as weaker instruction-focused models benefit substantially from collaboration.

In the DeepSeek family, the general-purpose \texttt{DeepSeek-V3.2} consistently outperforms the reasoning-specialized \texttt{DeepSeek-R1}, reinforcing the idea that scale and training factors may outweigh explicit reasoning design.

Overall, model family and specialization shape individual performance, but team-based aggregation reduces these differences by stabilizing weaker models.

\section{Discussion}
\label{discussion}

\subsection{Diversity of Model Answers and Team Performance}
\label{diversity}

In the following analysis, we focus on team configurations where \texttt{Qwen3-235B-A22B-Thinking}, the strongest individual model and the captain of the highest-performing team overall, serves as the captain.

\begin{figure}[t]
  \centering
  \includegraphics[width=0.8\linewidth]{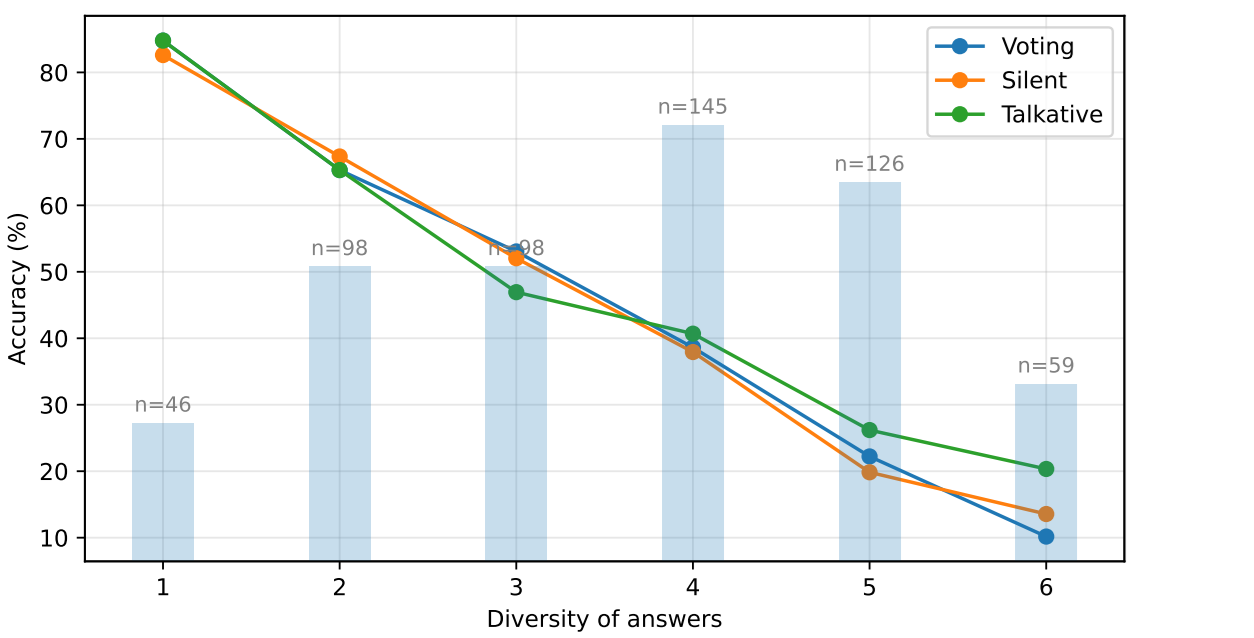}
  \caption{Team accuracy as a function of answer diversity $d$. Shaded bars indicate the number of questions ($n$) corresponding to each value of $d$.}
  \label{fig:answer_diversity}
\end{figure}

We study how answer diversity affects team performance. For each question, diversity is defined as the number of distinct answers produced by the six models ($d \in \{1,\dots,6\}$). Here, $d=1$ means full agreement and $d=6$ complete disagreement. Figure~\ref{fig:answer_diversity} shows team accuracy as a function of $d$.

Across all strategies, accuracy decreases sharply as diversity increases. When models agree ($d=1$), performance exceeds 80\%. Under maximal disagreement ($d=6$), it falls below 25\%.

\begin{figure}[t]
\centering
\includegraphics[width=1.0\linewidth]{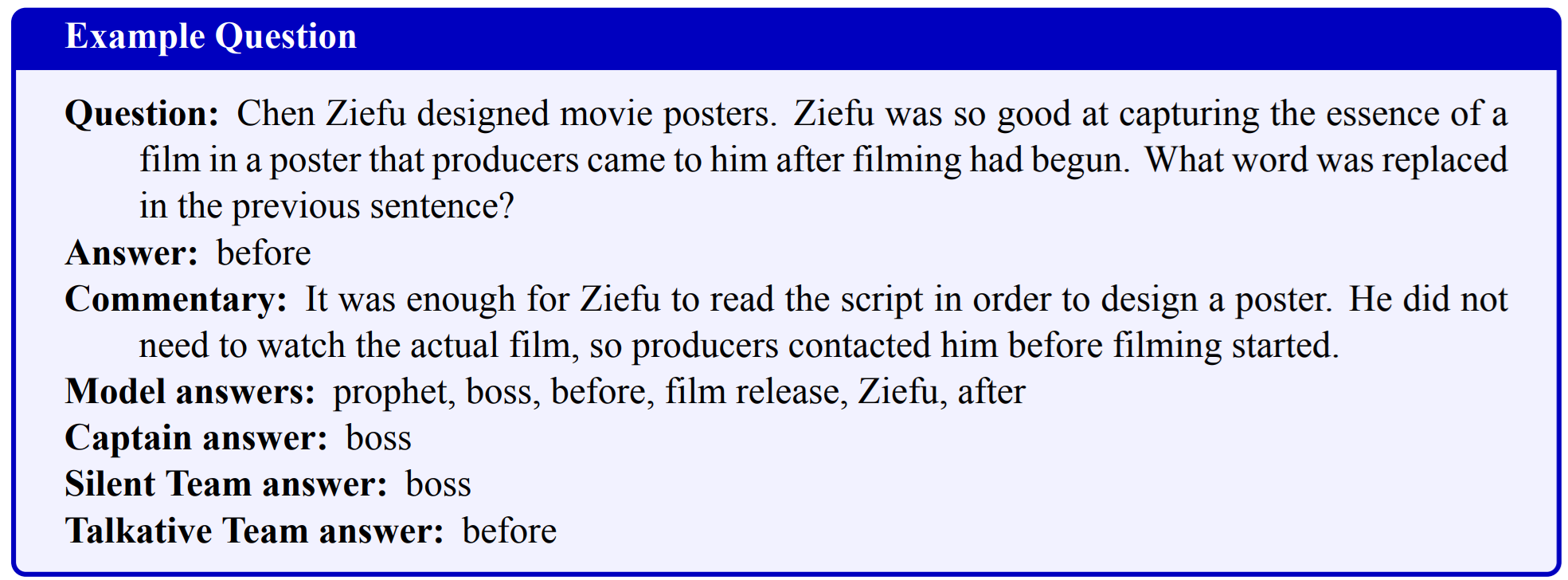}
\caption{Example of a high-diversity question ($d=6$) with a correct Talkative Team answer.}
\label{fig:example_question}
\end{figure}

The Talkative Team behaves differently in high-disagreement cases. For $d=5$ and $d=6$, it consistently outperforms Voting and Silent Team. This suggests that short explanations help the captain interpret conflicting answers and recover useful signals even without consensus (see Figure~\ref{fig:example_question}). 

However, at moderate disagreement ($d=3$), Talkative slightly underperforms the other methods. One possible explanation is that partially conflicting explanations introduce additional noise and complicate decision-making.

A natural question is whether diversity simply reflects question difficulty. Harder questions might lead to more disagreement, while easier ones may produce consensus. To test this, we analyze the 439 questions with available human statistics.

Across these questions, human teams achieve an average per-question solve rate of 49.83\%, while the best LLM configuration (Talkative Team) achieves 47.61\% accuracy.

The Spearman correlation between diversity $d$ and human success rate is weak ($\rho=-0.0914$) and only marginally non-significant ($p=0.055$). This suggests that disagreement is not merely a proxy for overall difficulty.

\begin{figure}[t]
  \centering
  \includegraphics[width=1.0\linewidth]{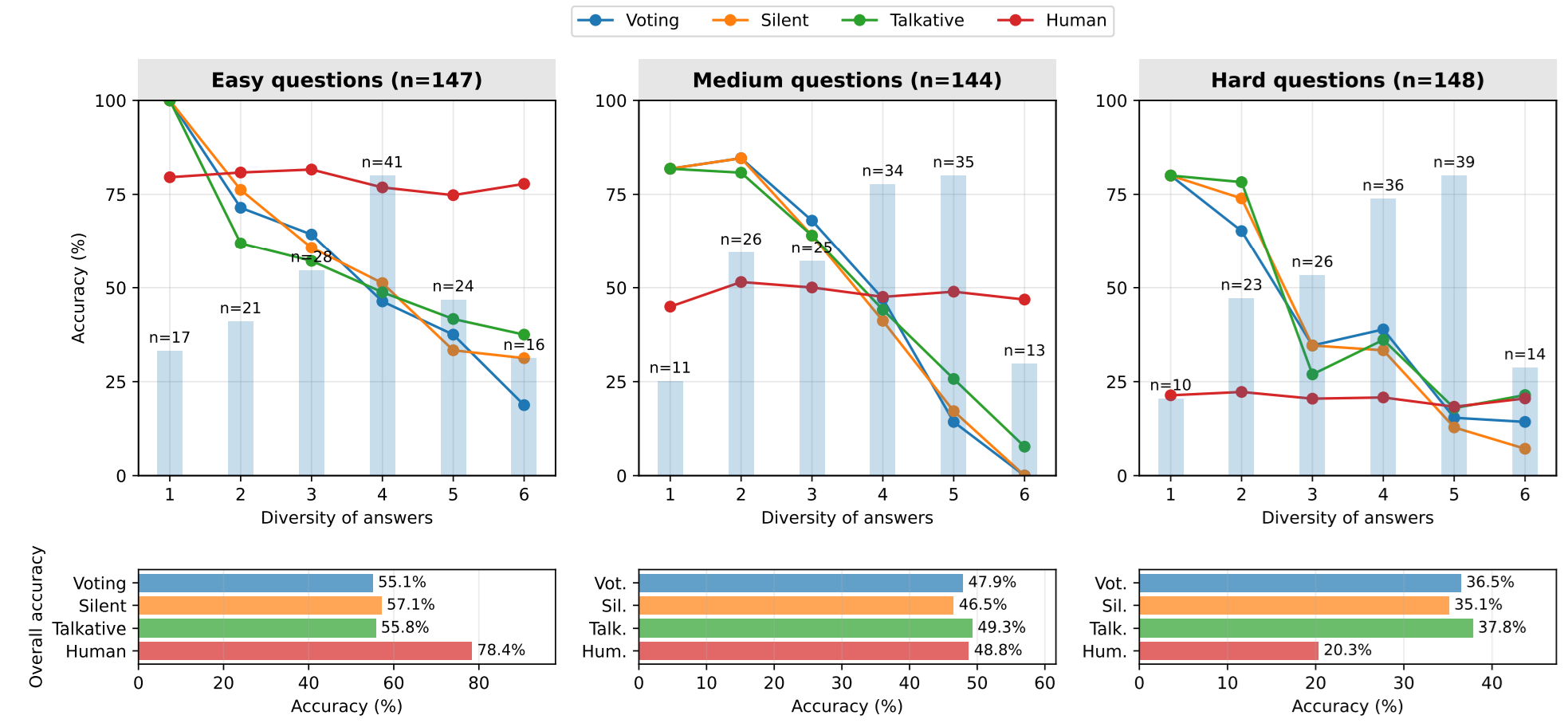}
  \caption{Team and human accuracy as a function of answer diversity across different difficulty levels. }
  \label{fig:answer_diversity_by_complexity}
\end{figure}

To explore this further, we divide the 439 questions into Easy, Medium, and Hard groups based on human performance (Figure~\ref{fig:answer_diversity_by_complexity}). In all three groups, higher diversity remains associated with lower accuracy. Thus, disagreement reflects instance-level uncertainty more than coarse difficulty categories.

Importantly, for $d \geq 5$, Talkative Team consistently outperforms the other strategies across all difficulty levels. The benefit of explanatory communication therefore appears robust and most pronounced under strong inter-model conflict.

Overall, inter-model disagreement serves as a useful signal for coordination. Simple aggregation works well under low disagreement, whereas communication-intensive strategies are most beneficial under high uncertainty. This suggests that future systems could adapt their coordination strategy dynamically based on observed diversity.

\subsection{Self-Preference Bias}

We examine whether the captain systematically favors its own initial answer and whether this behavior leads to errors.

In Talkative Team, the captain selects its own answer in 41.08\% of cases; in Silent Team -- 47.55\% (Table~\ref{tab:SelfPreference}). In both settings, self-selection is associated with higher accuracy. In Talkative Team, accuracy is 62.13\% when the captain keeps its own answer, compared to 31.75\% when it switches. A similar pattern holds in Silent Team (51.47\% vs.\ 34.33\%).

\begin{table}[t]
\begin{center}
\begin{tabular}{|l|rrrr|}
\hline
\bf Strategy & \bf Self-choice (\%) & \bf Acc (Self) (\%) & \bf Acc (Not Self) (\%) & \bf Self $\cap$ Majority (\%) \\
\hline
Silent Team    & 47.55 & 51.47 & 34.33 & 80.15 \\
Talkative Team & 41.08 & 62.13 & 31.75 & 83.40 \\
\hline
\end{tabular}
\end{center}
\caption{Self-preference behavior and conditional accuracy of the captain in Silent and Talkative teams. Self-choice denotes the proportion of cases in which the captain selected its own initial answer.}
\label{tab:SelfPreference}
\end{table}

Self-selection largely overlaps with majority agreement: in over 80\% of such cases, the captain’s initial answer matches the most frequent answer in the group. This suggests that apparent self-preference often reflects alignment with the majority rather than disregard for alternative views.

Although the captain is allowed to propose a new answer (see Appendix~\ref{app:prompt_talkative_team}), this never occurred in practice. All final decisions correspond to one of the initial model responses.

The two strategies differ in how self-selection is used. In Talkative Team, the captain relies on its own answer less often, but these decisions are more accurate. This indicates that access to peer rationales leads to more selective and better-calibrated confidence.

Overall, self-preference does not appear to be a systematic source of error. Instead, it reflects calibrated reliance on one’s initial judgment, especially when supported by group agreement. Communication reduces the frequency of self-selection while increasing its reliability.

\section{Conclusion}
\label{conclusion}

We examined whether teams of large language models can approximate aspects of human collective intelligence in the \textit{What? Where? When?} game. Using several interaction strategies and a newly constructed dataset of recent questions, we studied how diversity, communication, and coordination affect performance.

Across all model families, team-based aggregation consistently outperformed single-model baselines. In every setting, the team achieved higher accuracy than the captain model alone, indicating that collective decision-making provides systematic benefits beyond the strength of any individual model.

At the same time, we observe clear limitations. Although captains were allowed to generate new answers, they never proposed solutions outside the initial set of model responses. Final decisions were always selections among existing candidates. This suggests that current LLM teams primarily act as selection and error-filtering mechanisms rather than systems capable of genuine collective creativity.

Our results also show that coordination matters. Explanatory communication is especially helpful under high disagreement, while simple voting remains competitive when disagreement is low. This points to the potential of adaptive systems that adjust their interaction strategy based on observed uncertainty.

Several directions remain open. Future work could explore dynamic role assignment, multi-round deliberation, explicit hypothesis generation, and more complex coordination protocols such as hierarchical or multi-stage aggregation strategies. Extending evaluation to other domains with indirect reasoning and ambiguity would further test the generality of these findings.

Overall, LLM teams demonstrate meaningful forms of collective intelligence, but their behavior remains largely limited to aggregating individual outputs. Enabling deeper collaborative reasoning remains an important challenge for future multi-agent language systems.

\bibliography{paper.bib}
\bibliographystyle{dialogue}

\appendix

\section{Prompts}
\label{app:prompts}

This appendix provides the full text of the prompts used in our experiments. Placeholders such as \texttt{\{question\}} and \texttt{\{answer1\}} denote fields populated at runtime. All experiments were conducted in Russian, which is the original language of the dataset. For readability, the prompts in the appendix are shown in English translation.

\subsection{Prompt for generating individual answers}
\label{app:prompt_individual_answer}

\begin{tcolorbox}[colback=gray!3!white, colframe=gray!60!black, width=\linewidth, fonttitle=\bfseries]
\begin{lstlisting}[basicstyle=\ttfamily\small, breaklines=true]
You are participating in an intellectual game. Please briefly reason about the following question and give the correct answer.
Question: {question}. 
Output the reasoning and answer in JSON format: 
{
  "reasoning": "your reasoning here", 
  "answer": "your answer here"
}
\end{lstlisting}
\end{tcolorbox}

\subsection{Gemini prompt for answer normalizing}
\label{app:prompt_gemini_normalize_answers}

\begin{tcolorbox}[colback=gray!3!white, colframe=gray!60!black, width=\linewidth, fonttitle=\bfseries]
\begin{lstlisting}[basicstyle=\ttfamily\small, breaklines=true]
You are an expert evaluating responses in an intellectual game. Your task is to assess the diversity of answers to the questions.
For each question, you are given: 
  "id" - the question identifier, 
  "question" - the question itself, 
  "correct_answer" - the correct answer, 
  "variations" - acceptable answer variations that are counted as correct, 
  "answer1", ... "answer6" - the answers you need to evaluate, 
Return only the JSON without additional comments, where each answer is a dictionary with the keys: 
  "id" - the question identifier, 
  "answer variants" - a dictionary where the keys are strings like "variant 1", "variant 2", etc., and the values are lists of answers that are not semantically different from each other or are simply identical.
Do not evaluate the correctness of the answers, but their semantic diversity as responses to the given question.
All six answers must be placed into one of the lists. Lists may contain repetitions.
Check that you have evaluated all {answer_count} answers from the list.
List of questions and answers: {answers}
\end{lstlisting}
\end{tcolorbox}

\subsection{Silent team prompt}
\label{app:prompt_silent_team}

\begin{tcolorbox}[colback=gray!3!white, colframe=gray!60!black, width=\linewidth, fonttitle=\bfseries]
\begin{lstlisting}[basicstyle=\ttfamily\small, breaklines=true]
You are participating in an intellectual game. You are given the following question and six answer variants. It is unknown whether any of the variants is correct.
  Question: {question}.
  Answer 1: {answer1}. 
  ...
  Answer 6: {answer6}. 
Briefly review the answer variants.
If the correct answer is listed, select it.
If none of the suggested variants is correct, suggest your own correct answer.
Provide the result in JSON format:
{
  "reasoning": "your brief thoughts are here",
  "answer": "your answer here"
}
\end{lstlisting}
\end{tcolorbox}

\subsection{Talkative team prompt}
\label{app:prompt_talkative_team}

\begin{tcolorbox}[colback=gray!3!white, colframe=gray!60!black, width=\linewidth, fonttitle=\bfseries]
\begin{lstlisting}[basicstyle=\ttfamily\small, breaklines=true]
You are participating in an intellectual game. You are given the following question and six answer variants. For each answer, the player's reasoning is provided. It is unknown whether any of the variants is correct.
  Question: {question}.
  Answer 1:
  {
    "reasoning": {reasoning1},
    "answer": {answer1}
  },
  ...
  Answer 6:
  {
    "reasoning": {reasoning6},
    "answer": {answer6}
  }
Briefly review the answer variants and their explanations. If the correct answer is listed, select it. If none of the suggested variants is correct, suggest your own correct answer.
Provide the result in JSON format:
{
  "reasoning": "your brief thoughts are here",
  "answer": "your answer is here"
}
\end{lstlisting}
\end{tcolorbox}

\subsection{LLM-as-a-Judge prompt}
\label{app:prompt_llm_as_a_judge}

\begin{tcolorbox}[colback=gray!3!white, colframe=gray!60!black, width=\linewidth, fonttitle=\bfseries]
\begin{lstlisting}[basicstyle=\ttfamily\small, breaklines=true]
You are an expert, a researcher of answers in an intellectual game. Your task is to evaluate the answers to the questions. Each answer is evaluated separately, independently of other answers.
For each question, you are given:
  "id" - question ID,
  "question" - the question,
 "answer" - the answer you have to evaluate,
  "correct_answer" - the correct answer,
  "variations" - acceptable answer variants that are also counted as correct.
Return without further comments only a JSON list of ratings, where each rating is a dictionary with keys:
  "id" - question ID,
  "is_correct" - Boolean value indicating whether the answer is correct.
\end{lstlisting}
\end{tcolorbox}

\end{document}